\documentclass[journal]{IEEEtran}

\ifCLASSINFOpdf
\else
\fi

\usepackage[pdftex]{graphicx}
\usepackage{graphicx,bm} 
\usepackage{caption,subcaption,tabularx}
\usepackage[numbers,sort&compress]{natbib}
\usepackage{multirow}
\usepackage{xcolor}
\usepackage[symbol]{footmisc}

\usepackage{hyperref}
\begin{document}
%
\title{Beyond PRNU: Learning Robust Device-Specific Fingerprint for Source Camera Identification}
%
%
%

\author{Manisha, Chang-Tsun Li$^{\ast\ast}$,~\IEEEmembership{Senior Member,~IEEE,}
        Xufeng Lin,
        and Karunakar A. Kotegar,~\IEEEmembership{Senior Member,~IEEE} 
\thanks{Manisha and Karunakar A. Kotegar are with the Department
of Computer Applications, Manipal Institute of Technology, Manipal Academy of Higher Education, Manipal, India (e-mail: nimamanisha@gmail.com, karunakar.ak@manipal.edu).}
\thanks{Xufeng Lin and Chang-Tsun Li are with the School of Information Technology, Deakin University, Waurn Ponds Campus, Geelong, Australia (e-mail: \{xufeng.lin,changtsun.li\}@deakin.edu.au).}
\thanks{$^{\ast\ast}$Corresponding author: Chang-Tsun Li (changtsun.li@deakin.edu.au).}
}


\maketitle

\begin{abstract}
Source camera identification tools assist image forensic investigators to associate an image in question with a suspect camera. Various techniques have been developed based on the analysis of the subtle traces left in the images during the acquisition. The Photo Response Non Uniformity (PRNU) noise pattern caused by sensor imperfections has been proven to be an effective way to identify the source camera. The existing image forensic literature suggests that the PRNU is the only fingerprint that is device-specific and capable of identifying the exact source device. However, the PRNU is susceptible to camera settings, image content (e.g., scene details), image processing operations (e.g., simple low-pass filtering or JPEG compression), and counter-forensic attacks. A forensic investigator unaware of malicious counter-forensic attacks or incidental image manipulations is at the risk of getting misled. The spatial-synchronization requirement during the matching of two PRNUs also represents a major limitation of the PRNU. In recent years, deep learning based data-driven approaches have been successful in identifying source camera models. However, the identification of individual cameras of the same model through these data-driven approaches remains unsatisfactory. In this paper, we bring to light the existence of a new robust data-driven device-specific fingerprint in digital images which is capable of identifying the individual cameras of the same model. It is discovered that the new device fingerprint is location-independent, stochastic, and globally available, which resolve the spatial synchronization issue. Unlike the PRNU, which resides in the high-frequency band, the new device fingerprint is extracted from the low and mid-frequency bands, which resolves the fragility issue that the PRNU is unable to contend with. Our experiments on various datasets also demonstrate that the new fingerprint is highly resilient to image manipulations such as rotation, gamma correction, and aggressive JPEG compression.
\end{abstract}


\IEEEpeerreviewmaketitle

\section{Introduction}
\label{intro}
\IEEEPARstart{T}{he} availability of cost-effective smartphones has made the creation of digital images easier than ever before. As a result, digital images have become ubiquitous. Sometimes, images are used for malicious purposes such as fake news creation, pedo-pornography, and violence instigation. In such cases, investigators may be interested in identifying the camera that has been used to capture the image in question. However, one can easily manipulate the image using freely accessible photo editing tools without leaving any visually detectable traces. This reduces the credibility of the digital images to serve as valid evidence in the court of law. In such a crime scene investigation, source camera identification tools assist the forensic investigator to ensure the trustworthiness of the digital image in question by identifying the origin of the image. Albeit the fact that the information about the camera model, date and time are available in the image header file (i.e., the metadata or EXIF data), it cannot be used for forensic purposes as they can be easily tampered with. As a result, blind approaches have been developed by investigating the self-contained image data to identify the source camera, rather than relying on the auxiliary metadata. The blind source identification techniques take the advantage of the subtle traces left on the image by various modules involved in the image acquisition pipeline. These traces carry certain information that is unique to the source camera of the image and hence can be used as device fingerprint. 
\par Different camera models employ different lens systems to focus the light on the sensor, resulting in a lens distortion in the image \cite{san}. Each camera model has a unique lens distortion pattern that helps in identifying the camera model of the given image. Since the camera sensor can measure only one color at each pixel location, a Color Filter Array (CFA) is used so that the individual pixels only receive the light of a certain wavelength. The missing color information is estimated from the neighboring pixels through a process called demosaicing (interpolation). The camera models from different brands have their own demosaicing algorithms to reconstruct the missing color values. Thus, the inter-pixel dependencies caused by the CFA demosaicing have also been used to identify the camera model of the image \cite{swami,bay,cao,chen2015}. The cameras from different brands follow proprietary post-processing algorithms such as white balancing and JPEG compression. Hence, the statistical traces left on the image by the post-processing operations have been used to identify the camera model of the image as well \cite{deng, sor}.
\par However, the aforementioned intrinsic traces can only differentiate the cameras of different brands or models and cannot trace back the individual devices belonging to the same brand and model. Thus, it is necessary to extract the features that can identify the exact camera. The imperfections in the manufacturing of the camera sensor lead to variation in different pixels' sensitivity to the light of the same intensity. As a result, each camera leaves unique Sensor Pattern Noise (SPN) \cite{lukas} in every image that it has captured. The SPN has two major components: Fixed Pattern Noise (FPN) and Photo Response Non-Uniformity (PRNU). The FPN and PRNU depend on dark current and non-uniformity in the pixels of the camera sensor, respectively. The FPN caused by the dark current in the camera sensor is usually suppressed using the dark frame within the camera. Therefore, PRNU is the only dominant component of the SPN that is present in the output image. The PRNU noise pattern being a deterministic component, is stable over time and remains approximately the same if multiple images of the same scene are captured \cite{lukas}. It is independent of scene detail in the image and depends only on the physical characteristics of the camera sensor. Thus, each camera is characterized by its unique PRNU, which can be used as a device fingerprint for differentiating individual cameras of the same model or brand \cite{lukas}. Due to these distinct characteristics, the PRNU has drawn much attention in the device-level source camera identification.
\par Over the past two decades, a variety of methods have been proposed to identify the source camera by extracting the PRNU from the images \cite{goljan, chen, lukas, aks}. Since the PRNU resides in the high-frequency band of the image, most of the source identification techniques \cite{lin, lin2015, marra,aks} employ the denoising filter proposed in \cite{lukas} to extract the PRNU from an image. However, the PRNU, a form of hand-crafted device fingerprint, can be attenuated by many factors, such as scene details \cite{Li}, PRNU filtering \cite{lin}, periodical image processing operations \cite{lin2015}, cameras settings \cite{quan21}, etc., and can be easily removed to hinder source identification through simple low-pass filtering \cite{adap, li2009}. Moreover, the extracted PRNU pattern has the same pixel number as the original image, which incurs high computational and storage costs for source-oriented image clustering \cite{lin2016, ame2014, li2017}. To overcome the computational complexity, the PRNU is either extracted from a small portion of the image or formated as a more compact representation  \cite{lin2016, li2018}. However, this results in the loss of important features characterizing the source camera, thus compromising the accuracy of source camera identification. 
\par Although PRNU fingerprint has been proven to be an effective way to identify the source camera, it may be undesirable for anti-forensic or criminal attackers such as pedophiles or fake news creators who want to retain their anonymity when sharing images. It would be desirable to unlink the images from their source camera when anti-forensics is needed. Thus, various counter forensic techniques have been developed that enabled eliminating or suppressing the PRNU from images to anonymize images. Also, the desynchronization operations such as rotation, cropping, resizing, and filtering prevent the detection of PRNU fingerprint by disturbing the relationship between the neighboring pixel and thereby hindering the correct identification of the source camera. There is currently a range of approaches in the literature \cite{adap, dirik, li2009, dirik2014} that are effective in defeating state-of-the-field source camera identification techniques by eliminating the PRNU fingerprint. A forensic investigator having no idea of a counter-forensic attack on the image would have a high probability of getting misled by the attacker.


\par In recent years, inspired by the success of deep learning in computer vision, image forensics researchers developed Convolutional Neural Networks (CNNs) based data-driven systems \cite{bondi, tuama, yao, fre, huang, wang, mani} to identify the source. The current state-of-the-art CNN-based approaches \cite{chen2017, yang2019, ding2019, sameer2020, mandelli} provide promising solutions to camera model identification. However, as we will discuss in Section \ref{lit}, they are not capable of effectively identifying devices of the same models or brands. Therefore, it is desirable to have a new data-driven method for extracting device fingerprint that is capable of differentiating individual devices of the same model. The major contributions of the proposed work are as follows. 
\begin{enumerate}
	\item We make evident the presence of a new \textit{non-PRNU} device-specific fingerprint in digital images and develop a data-driven approach using CNN to extract the fingerprint. 
	\item We show that the new device-specific fingerprint is mainly embedded in the low- and mid-frequency components of the image, hence is more robust than the fingerprint (e.g., PRNU) residing in the high-frequency band. 
	\item We show that the \textit{global, stochastic, and location-independent} characteristics of the new device fingerprint make it a robust fingerprint for forensic applications.
	\item We validate the robustness of the new fingerprint on common image processing operations. 
\end{enumerate}
\par The remainder of the paper is structured as follows. Section \ref{lit} reviews related works. Section \ref{meth} presents the proposed data-driven approach to explore the presence of a new fingerprint. Experimental results and discussions are given in Section \ref{results}. Finally, conclusions are drawn in Section \ref{con}.

\section{Literature Review}
\label{lit}
In light of the high sensitivity of PRNU (i.e., a form of hand-crafted device fingerprint) to common image processing (e.g., low pass filtering and compression) in source device identification (SDI), the past few years have witnessed the successful attempts of taking a data-driven approach based on CNN \cite{bondi, tuama, yao, fre, huang, wang}. Bondi et al., \cite{bondi} proposed a model based on CNN and SVM for camera model identification. The method aimed at learning the camera model specific features directly from the image rather than depending on the hand-crafted features. The proposed method was able to work on small image patches (64$\times$64 pixels) with 93\% accuracy on 18 different camera models. Tuama et al., \cite{tuama} developed a CNN model by modifying the AlexNet. A pre-processing layer consisting of a high-pass filter was added to the CNN model to reduce the impact of scene details. The CNN was trained on image patches of size 256$\times$256 pixels to identify the model of the source camera. A CNN-based robust multi-classifier was developed by Yao et al., \cite{yao} to identify the camera model. The method used 256 non-overlapping patches of size 64$\times$64 pixels extracted from the central portion of the image and achieved a classification accuracy of nearly 100\% over 25 camera models. Freire-Obregón et al., \cite{fre} proposed a source camera identification method for mobile devices based on deep learning. The method involved training the CNN model using 256 image patches of size 32$\times$32 pixels extracted from each image. The performance is degraded when the multiple cameras of the same brand and model are considered. Huang et al., \cite{huang} developed a CNN model and evaluated the effect of the number of convolutional layers on the performance of the model. The work suggested that the deeper CNN can achieve better classification accuracy. Also, the work showed that replacing the Softmax layer in the CNN with the SVM classifier will improve the identification accuracy. Wang et al., \cite{wang} developed a CNN model by modifying the AlexNet and equipping it with a Local Binary Pattern (LBP) pre-processing layer to allow the CNN to have more focus on the intrinsic source information (such as PRNU \cite{lukas}, lens distortion noise pattern \cite{san} and traces of color dependencies related to CFA interpolation \cite{swami}) that is concealed in the image rather than the scene details. The images were divided into non-overlapping patches of size 256$\times$256 pixels to train the CNN model. The proposed method achieved an identification accuracy of 98.78\% over 12 camera models. Albeit the encouraging fact that the afore-mentioned CNN-based methods are able to effectively identify the model of the source camera of the image in question, they are by no means capable of effectively differentiating individual cameras of the same model.
\par In \cite{chen2017}, Chen et al., proposed a method to investigate the task of brand, model and device identification using a Residual network. The model was trained by using one patch of size 256$\times$256 extracted from each image. Despite having the good ability to identify the brand and model, it was able to achieve only 45.81\% accuracy when different devices of the same brand and model were involved in the experiment. 
Yang et al., \cite{yang2019} developed a content-adaptive fusion network to identify the source devices of small image patches of size 64$\times$64. The exact device detection accuracy, in this case, was 70.19\% when only 3 devices of the same model were used. Ding et al., \cite{ding2019} further improved the task of exact device identification by developing a residual network with a multi-task learning strategy by taking advantage of both hand-crafted and data-driven technologies. However, the model was able to achieve only 52.4\% accuracy on 74 cameras from the Dresden dataset and 84.3\% accuracy on 51 cameras from the Cellphone dataset, on image patches of size 48$\times$48 pixels.  
\par In almost all the existing techniques, a large number of images are used to train the CNN models for source identification. However, collecting a large number of images is infeasible in realistic scenarios. Thus, Sameer et al., \cite{sameer2020} addressed the task of source camera identification, with a limited set of images per camera. A deep Siamese network was trained using a few shot learning technique by considering pairs of 64$\times$64 pixel image patches extracted from 10 images per camera. The method outperformed other state-of-the-art techniques, with a few training samples. Mandelli et al., \cite{mandelli} proposed a 2-channel based CNN that learns a way of comparing image noise residual and camera fingerprint at the patch level. Also, the method considered the scenario in which PRNU fingerprint and images are geometrically synchronized. The method was effective in achieving the enhanced source identification accuracy. However, it depends on the PRNU fingerprint for SDI. Hence, it works only when there is no pixel misalignment between the PRNU fingerprint or when the PRNU is not attacked.
\par Although many progresses in differentiating individual devices \cite{chen2017, yang2019, ding2019, sameer2020, mandelli} using the data-driven approach have been made to address PRNU's fragility issue, the accuracy is still far from forensically satisfactory. Thus, exact SDI remains a challenging task. Furthermore, the existing CNN models for source identification work only when there is no pixel misalignment is present in the images. However, image manipulations and counter forensic attacks can defeat the state-of-the-art source camera identification techniques by breaking the structural relationship between the neighboring pixels and eliminating the PRNU noise embedded in the image. Moreover, such a spatial synchronization requirement (i.e., all image patches need to be taken from the same location) is in no way feasible if the original image has been subjected to cropping or affine transformations such as scaling, rotation, shearing, and translation, etc. In view of the vulnerabilities of PRNU to counter forensic attacks and common image processing techniques, the research question then arises: \textbf{is there some form of robust device-specific fingerprint other than PRNU in the image that is capable of differentiating individual devices of the same model?} To answer this question, in this work we propose two simple yet effective methods, \emph{down sampling} and \emph{random sampling}, to generate PRNU-free images for learning robust device-specific fingerprint. By leveraging the generated PRNU-free images and the powerful learning capability of deep residual neural networks, we are able to establish the presence of robust device-specific fingerprint other than PRNU that can be used to identify the exact source device of images.

\section{Methodology}
\label{meth}
As we aim for the proposed model to learn a robust device fingerprint, our approach is to avoid the high-frequency bands of images. This is also to serve the purposes of 1) proving that the new device fingerprint is PRNU-irrelevant (i.e., its effectiveness is not attributed to the PRNU) and 2) enhancing the new device fingerprint’s immunity to the interference of strong scene details appearing in the high-frequency bands. Therefore, we present two different ways of creating PRNU-free image patches in Section III-A and III-B to reveal different characteristics of the new device fingerprint and to align the patch formation methods naturally with the input requirement of the proposed deep learning model (i.e., III-C). The two ideas of forming the PRNU-free image patches are applicable to other learning models.

\subsection{Generation of PRNU-free Images with Down Sampling}

\label{methA}
\begin{figure}[!t]
\centering
\includegraphics[width=3.3in]{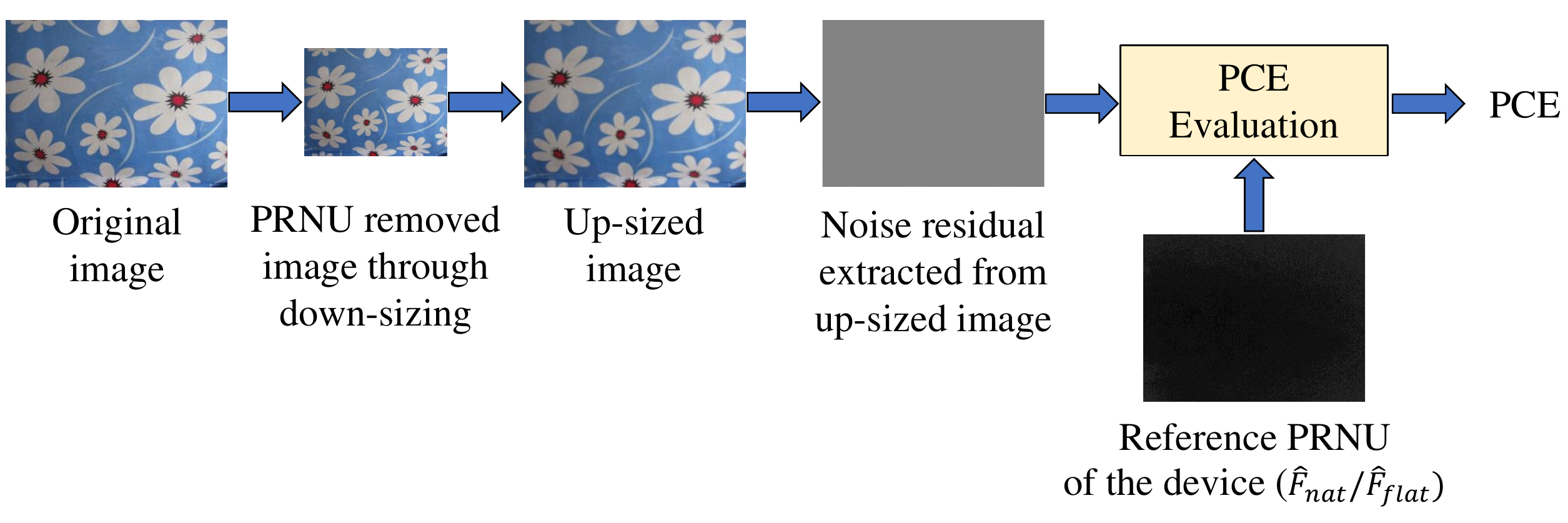}
\caption{PRNU fingerprint removal through down sampling and removal effectiveness evaluation based on PCE.}
\label{fig1}
\end{figure}

The PRNU fingerprint is embedded in the high-frequency bands of images \cite{lukas}. Hence, the removal of the PRNU fingerprint can be achieved by eliminating the high-frequency components of the image. There are various filtering operations for removing high-frequency components of the images. Since we will use ResNet as the fingerprint extractor (see Section III-C), which require patches of 224 x 224 pixels as input, our first method for PRNU removal is down sampling (a form of low-pass filtering), which serves not only the purpose of removing the PRNU but also conforming to the ResNet architectural requirement. We apply the bilinear filtering (low-pass filtering) to down-sample each image to 224$\times$224 pixels with 3 color channels. To ensure effective PRNU removal, we evaluate the similarity between the noise residual extracted from the PRNU-free image and the reference PRNU of its source camera using the widely adopted Peak to Correlation Energy (PCE) \cite{goljan}, as depicted in Fig. \ref{fig1}. The experiments in \cite{goljan} suggested a PCE detection threshold of 50 to determine whether two images are from the same source device or not. The greater the PCE value is, the higher the likelihood that the two images are of the same origin. Also, the counter-forensic attacks used to impede PRNU often fixed the PCE detection threshold at 50 to ensure effective PRNU removal and thereby unlinking the images from their source camera \cite{adap, dirik}. Thus, in the proposed work, we also use 50 as the threshold to determine whether the down-sampled images have become PRNU-free or not.
\par The PCE evaluation requires the noise residual that serves as the PRNU fingerprint of the image ($I_D$) and the reference PRNU fingerprint of its source device ($D$). We use the wavelet-based denoising method \cite{goljan} for noise residual extraction and construct the reference PRNU ($\hat{F}_{nat}$) from the noise residuals extracted from 50 natural images captured by the same device. To calculate the PCE value, we up-sample the down-sampled PRNU-free images to the same size as the reference PRNU for the evaluation of PCE. We also consider the reference PRNU ($\hat{F}_{flat}$) generated using 50 flat-field images. The flat-field images have fewer intensity variations which result in the generation of near-perfect camera fingerprint, making the PCE estimate more convincing. As we will see in Section \ref{resB}, the down sampling operation results in PCE values much lower than the threshold 50, indicating the effectiveness of PRNU removal in the down-sampled image $\hat{I}_D$.


\begin{figure*}[!t]
\centering
\includegraphics[width=7in]{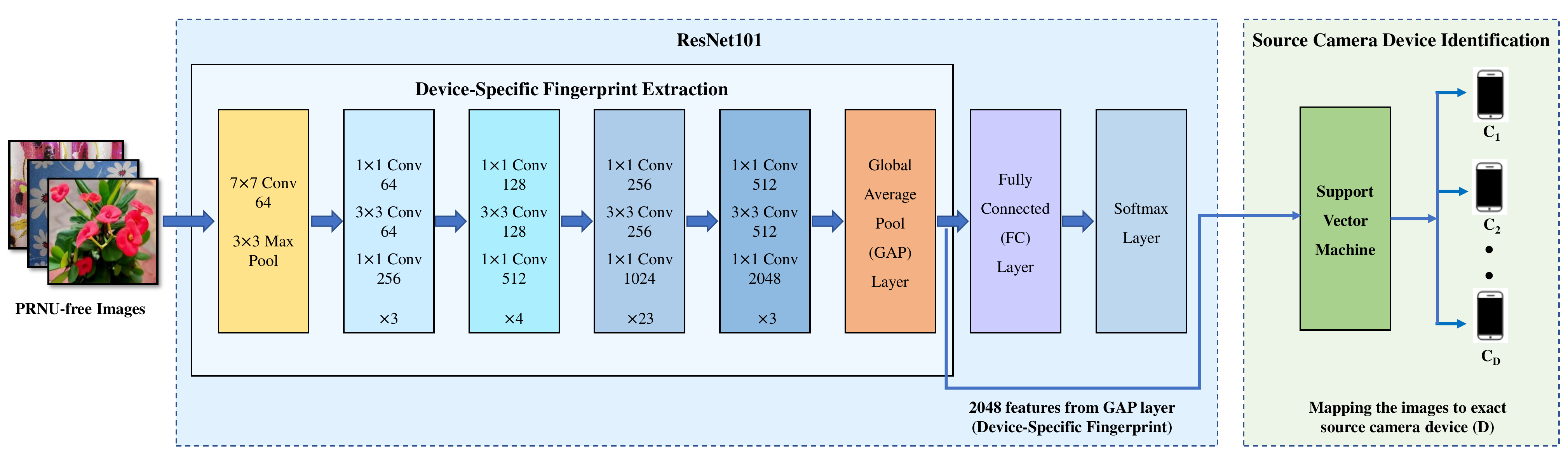}
\caption{The hybrid ResNet-SVM to extract the new device-specific fingerprint other than PRNU and to identify the source camera.}
\label{fig2}
\end{figure*}
`
\subsection{Generation of PRNU-free Images with Random Sampling}
\label{methD}
The patch formation approach based on down-sampling, can help our search for a device fingerprint in the low and mid-bands of images. However, it preserves the spatial relationship among the pixels within the patches and will not be able to address the spatial synchronization issue that the PRNU is unable to contend with. Therefore, in this second method, we propose to use random sampling to remove the PRNU fingerprint in images while complying with the ResNet architectural requirement. More importantly, it is intended to prove that there is a form of device fingerprint that is location-independent, stochastic, and globally present. Being a location-dependent and deterministic component \cite{lukas,chen}, the PRNU fingerprint requires proper spatial synchronization, thus forming image patches with pixels selected at random causes desynchronization and hence prevents the PRNU from contributing to source identification. Furthermore, the random sampling operation disrupts the contextual and semantic information of the image, which is irrelevant to the intrinsic fingerprint of the source device. By so doing, the model learns the device-specific fingerprint while ignoring such irrelevant but interfering information. For this purpose, we create patches of size 224$\times$224$\times$3 pixels by taking pixels from random locations either from the original image $I_D$ or the down-sampled image $\hat{I}_D$, thereby eliminating the structural relationship between the neighboring pixels. Here, each pixel is used only once while forming the patches. If a machine learning model is able to learn from the image patches formed by taking the pixels at random for device-level source camera identification, we make evident the global presence of location-independent, stochastic, and PRNU-irrelevant device-specific fingerprint that can be successfully used for image forensic applications. 


\subsection{Learning Device-Specific Fingerprint}
\label{methB}
With the PRNU-free images generated using the aforementioned two methods, we propose to use a hybrid system based on residual neural network (ResNet) \cite{res} and SVM classifier. ResNet is a variant of the CNN model with ``shortcut/skip connections'' that allows constructing substantially deeper networks for performance gain and has achieved state-of-the-art results on a range of computer vision benchmarks. The ResNet model serves as a feature extractor to extract device-specific fingerprint, which will then be used to train an SVM classifier to classify the images to their exact source cameras. 

The depth of the CNN is a crucial parameter \cite{yao, huang} that significantly affects the performance of the source camera identification task. However, building a deeper CNN is not as simple as stacking layers in the light of the vanishing gradient problem for neural networks relying on gradient-based training, e.g. back propagation. More specifically, a greater network depth makes it increasingly difficult to propagate the gradients back to the earlier layers of the network for weights update. As a consequence, the performance of the CNN saturates or even starts degrading rapidly as the CNN network goes deeper. Such an issue can be addressed by introducing skip connections between layers \cite{res}, which add the outputs of earlier layers to the deeper layers and thus allow uninterrupted gradient flow between layers in the network. This not only tackles the gradient vanishing problem but also enables the reusability of features at different layers \cite{he2016identity}, making it possible to learn both low-level and high-level features. Thus, our strategy is to make use of the deeper ResNet model to learn richer feature representation for extracting the device-specific fingerprint.

\par Feature/fingerprint extraction plays a major role in the success of source camera identification. It necessitates that extracted features exhibit discriminative characteristics among different cameras while remaining a high degree of homogeneity within the same class. The discriminative power of the features/fingerprints can be boosted by increasing the number of stacked layers. The ResNet model has many variants with a different number of layers such as ResNet18, ResNet34, ResNet50, ResNet101, etc. With the increase in the number of layers, the ResNet model can integrate multi-scale features from multiple layers to enhance the features discriminative power. In this work, we employ the ResNet model with 101 layers (ResNet101) as the backbone of our device fingerprint extractor. The proposed ResNet-based device-specific fingerprint extractor is shown on the left-hand side of Fig. \ref{fig2}.

\par In order to capture the device-specific fingerprint other than PRNU, we train the ResNet101 model using PRNU-free images as described in Section \ref{methA} and \ref{methD}. The number of neurons in the last fully connected (FC) layer is equal to the number of devices $D$ used for training the model. During the forward propagation, the PRNU-free images are passed through successive ResNet blocks and the GAP layer. Eventually, at the FC layer, the feature vector is reduced to the number of cameras $D$ to get $z=[z_1,z_2,....z_D]$. Finally, a Softmax function is applied to convert the scores produced by the FC layer to probabilities $(\hat{\bm{y}}=[\hat{y}_1,\hat{y}_2,....\hat{y}_D])$, where $\hat{y}_i$ is the probability of the image being taken with the $i^{th}$ camera:
\begin{equation}
\hat{y}_i = \frac{e^{z_i}}{\sum_{d=1}^D e^{z_d}}
\label{eq2}
\end{equation}
The prediction loss is then calculated based on the class distribution ($\hat{\bm{y}}$) predicted by the ResNet101 model and the one-hot encoded ground-truth class vector ($\bm{y}$) of the input image. For the proposed work, we make use of the \textit{categorical cross-entropy loss function} which is widely used for multi-class classification:
\begin{equation}
L(\hat{\bm{y}},\bm{y}) = -\sum_{i=1}^D y_i\log(\hat{y}_i)
\label{eq3}
\end{equation}
where $y_i$ and $\hat{y}_i$ are the actual class and the probability of the input image belonging to the $i^{th}$ camera, respectively. 

\subsection{Source Device Identification}
\label{methC}
To make it evident that the device fingerprint extracted by the ResNet101 model can serve the task of SDI, we use the multi-class SVM classifier with the Radial Basis Function (RBF) kernel to identify the source device for the PRNU-free image. We feed the images to the trained ResNet101 model and use the outputs from the GAP layer as the device-level fingerprints for training the SVM classifier. Once the SVM classifier is trained, it makes class predictions on the images in the test set based on the device fingerprint extracted from the GAP layer to perform the task of SDI. We use the same training set (75\% of the images of each camera) for training both the ResNet101 model and the SVM classifier, while the testing set (25\% of the images of each camera) is only used for testing the source camera identification performance of the SVM classifier, which ensures that the testing set is not involved in the training of feature extractor (i.e. the Resnet101 model).

\section{Experiments and Discussions}
\label{results}

\subsection{Experimental Setup}
\label{resA}
For the evaluation of the proposed method, we have used the images from the VISION dataset \cite{shul17}, Warwick Image Forensic Dataset \cite{warwick}, Daxing dataset \cite{daxing} and a custom-built dataset. All datasets have different characteristics in terms of camera model and brand, image content and number of devices from the same model. Because our work aims at identifying the device-specific fingerprint, we have considered only the images taken from different devices belonging to the same brand and model in these datasets. Details of the cameras and the images used are given in Table \ref{table1}. From the VISION dataset, in addition to natural images, we have also considered 50 flat-field images for the generation of the reference PRNU ($\hat{F}_{flat}$). The custom-built dataset includes images taken from two personal smartphone models. From each model, we have considered two different devices. Involving different datasets in the experiments ensures the diversity of the cameras and images used for the evaluation. For each dataset, the images are randomly divided into the training and testing sets, such that 75\% of the images from each device are randomly chosen for training the model and the rest 25\% for testing the model. We train the ResNet101 model for 20 epochs using the Adam optimizer with the mini-batch size and the learning rate set to 12 and 0.001, respectively. All experiments are conducted using Matlab2020b on an HP EliteDesk 800 G4 Workstation with an NVIDIA GeForce GTX 1080 GPU, 3.7 GHz Intel processor and 32GB RAM.

\begin{table}[!t]
\centering
\caption{Devices in different datasets used in our experiments}
\begin{tabular}{|c|c|c|c|c|}
\hline
\textbf{Dataset}                       & \textbf{Model}                                                      & \textbf{\begin{tabular}[c]{@{}c@{}}No. of \\ devices\end{tabular}} & \textbf{\begin{tabular}[c]{@{}c@{}}No. of images \\of each device\end{tabular}} & \textbf{\begin{tabular}[c]{@{}c@{}}Resolution\end{tabular}} \\ \hline
\multirow{4}{*}{\textbf{Vision}}       & iPhone 5c                                                           & 2                                                                  & 204                                                                      & 3264$\times$2448                                                            \\ \cline{2-5} 
                                       & iPhone 4s                                                           & 2                                                                  & 200                                                                      & 3264$\times$2448                                                            \\ \cline{2-5} 
                                       & iPhone 5                                                            & 1                                                                  & 204                                                                      & 3264$\times$2448                                                            \\ \cline{2-5} 
                                       & iPhone 6                                                            & 1                                                                  & 204                                                                      & 3264$\times$2448                                                            \\ \hline
\textbf{Warwick}                       & \begin{tabular}[c]{@{}c@{}}Fujifilm \\ X-A10\end{tabular}           & 2                                                                  & 197                                                                      & 3264$\times$2448                                                            \\ \hline
\multirow{6}{*}{\textbf{Daxing}}       & \begin{tabular}[c]{@{}c@{}}Huawei \\ P20\end{tabular}               & 2                                                                  & 190                                                                      & 2976$\times$3968                                                            \\ \cline{2-5} 
                                       & iPhone 6                                                            & 3                                                                  & 190                                                                      & 3264$\times$2448                                                            \\ \cline{2-5} 
                                       & iPhone 7                                                            & 3                                                                  & 190                                                                      & 4032$\times$3024                                                            \\ \cline{2-5} 
                                       & \begin{tabular}[c]{@{}c@{}}iPhone 8 \\ Plus\end{tabular}            & 1                                                                  & 190                                                                      & 4032$\times$3024                                                            \\ \cline{2-5} 
                                       & Oppo R9                                                             & 2                                                                  & 190                                                                      & 2304$\times$4096                                                            \\ \cline{2-5} 
                                       & Xiaomi 4A                                                           & 5                                                                  & 190                                                                      & 3120$\times$4160                                                            \\ \hline
\multirow{2}{*}{\textbf{Custom-built}} & \begin{tabular}[c]{@{}c@{}}Redmi \\ Note 3\end{tabular}             & 2                                                                  & 200                                                                      & 4608$\times$3456                                                            \\ \cline{2-5} 
                                       & \begin{tabular}[c]{@{}c@{}}Asus \\ Zenfone \\ MaxProM1\end{tabular} & 2                                                                  & 250                                                                      & 4608$\times$3456                                                           \\ \hline
\end{tabular}
\label{table1}
\end{table}

\begin{figure*}[!t]
    \centering 
\begin{subfigure}{0.25\textwidth}
  \includegraphics[width=\linewidth]{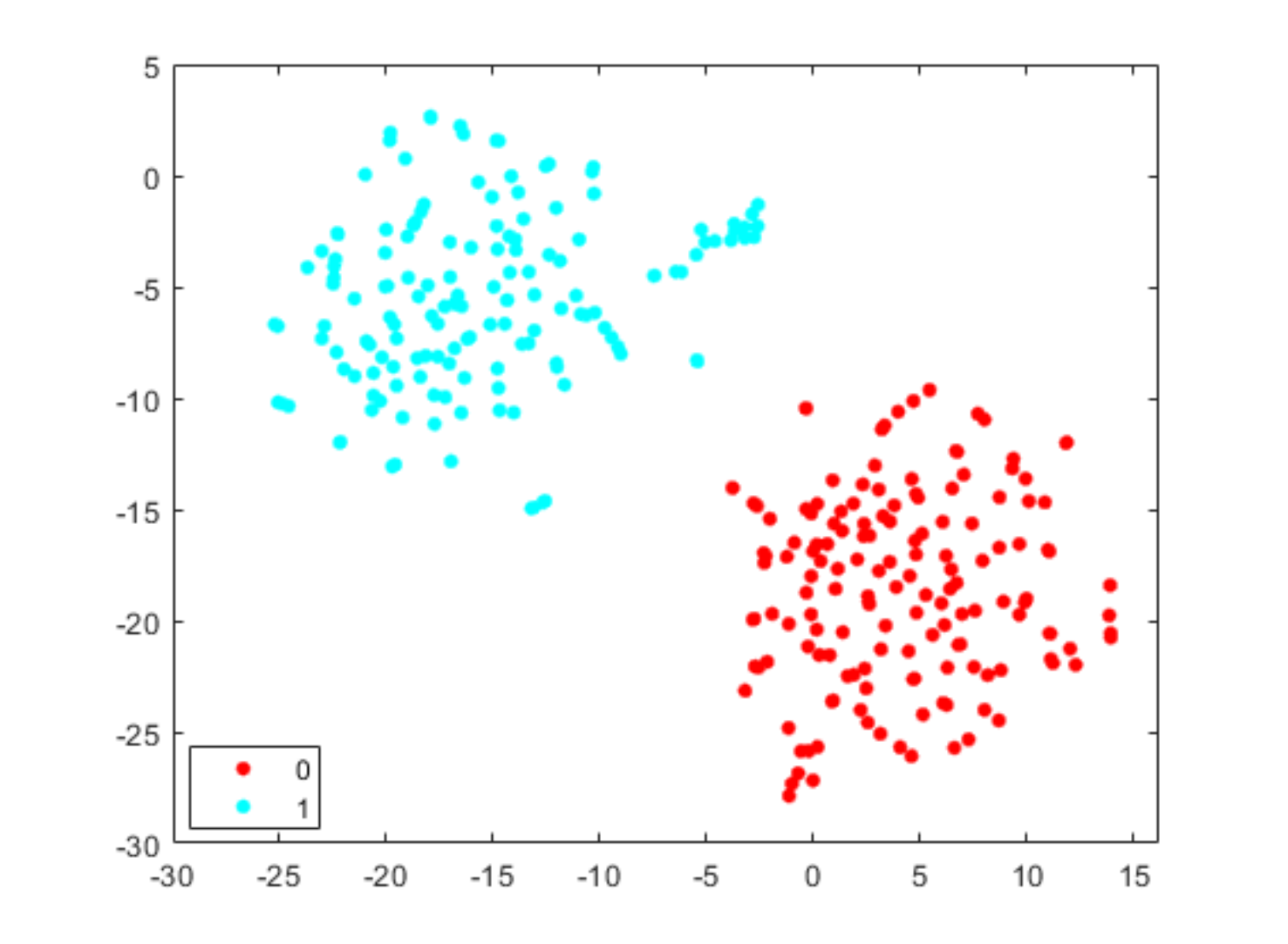}
  \centerline{(a)}
\end{subfigure}\hfil 
\begin{subfigure}{0.25\textwidth}
  \includegraphics[width=\linewidth]{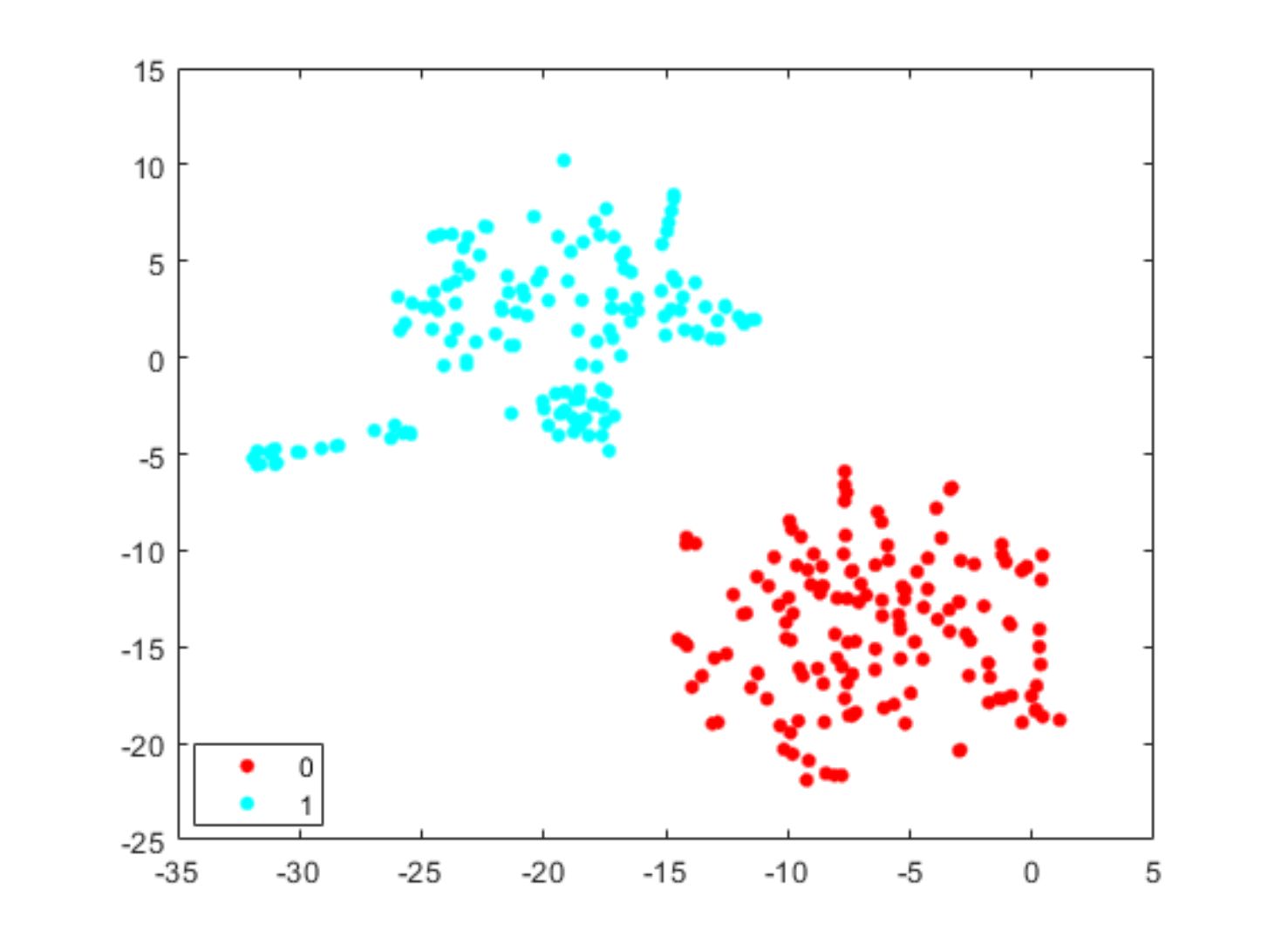}
  \centerline{(b)}
\end{subfigure}\hfil 
\begin{subfigure}{0.25\textwidth}
  \includegraphics[width=\linewidth]{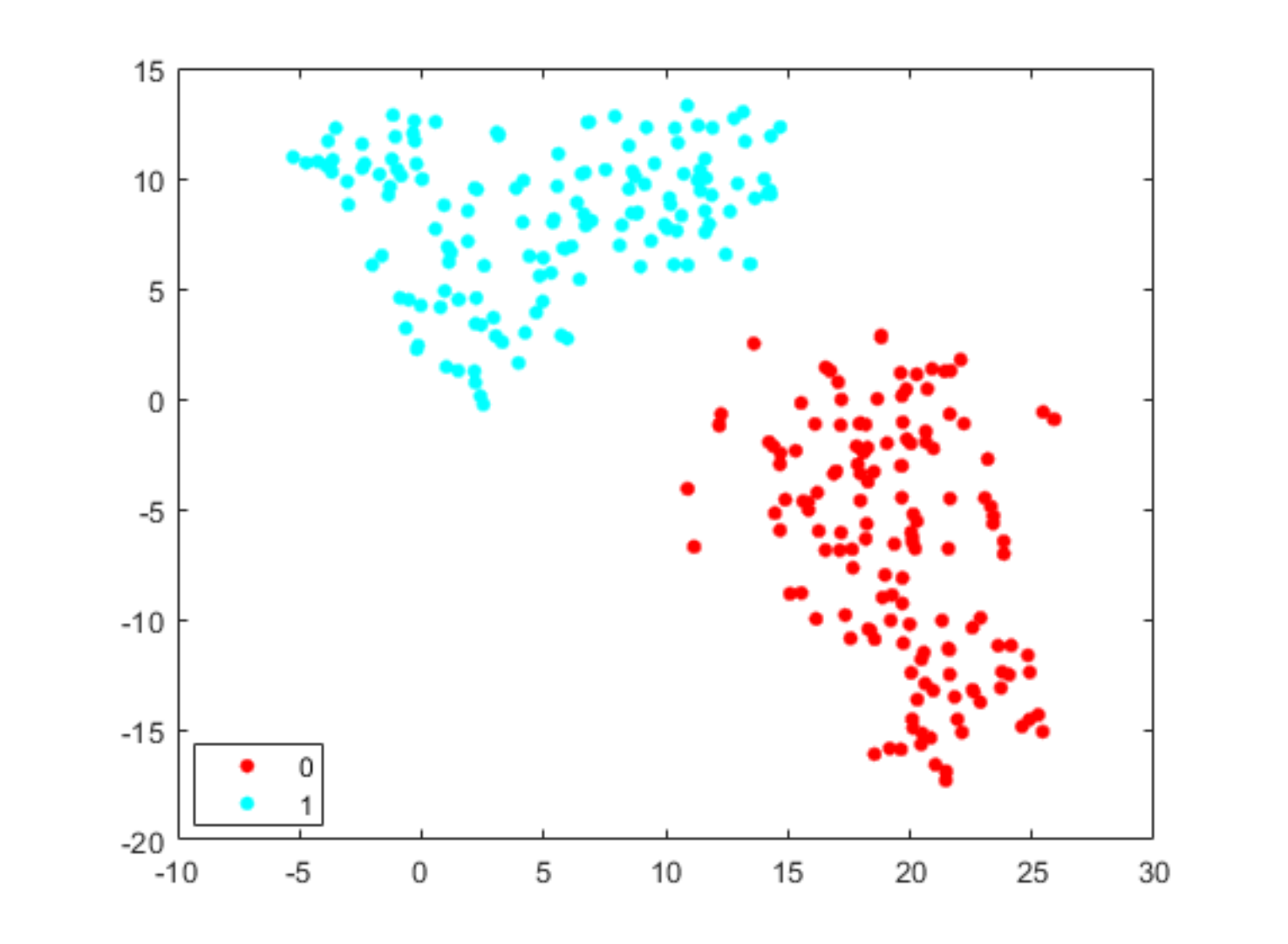}
  \centerline{(c)}
\end{subfigure} 
\medskip
\begin{subfigure}{0.25\textwidth}
  \includegraphics[width=\linewidth]{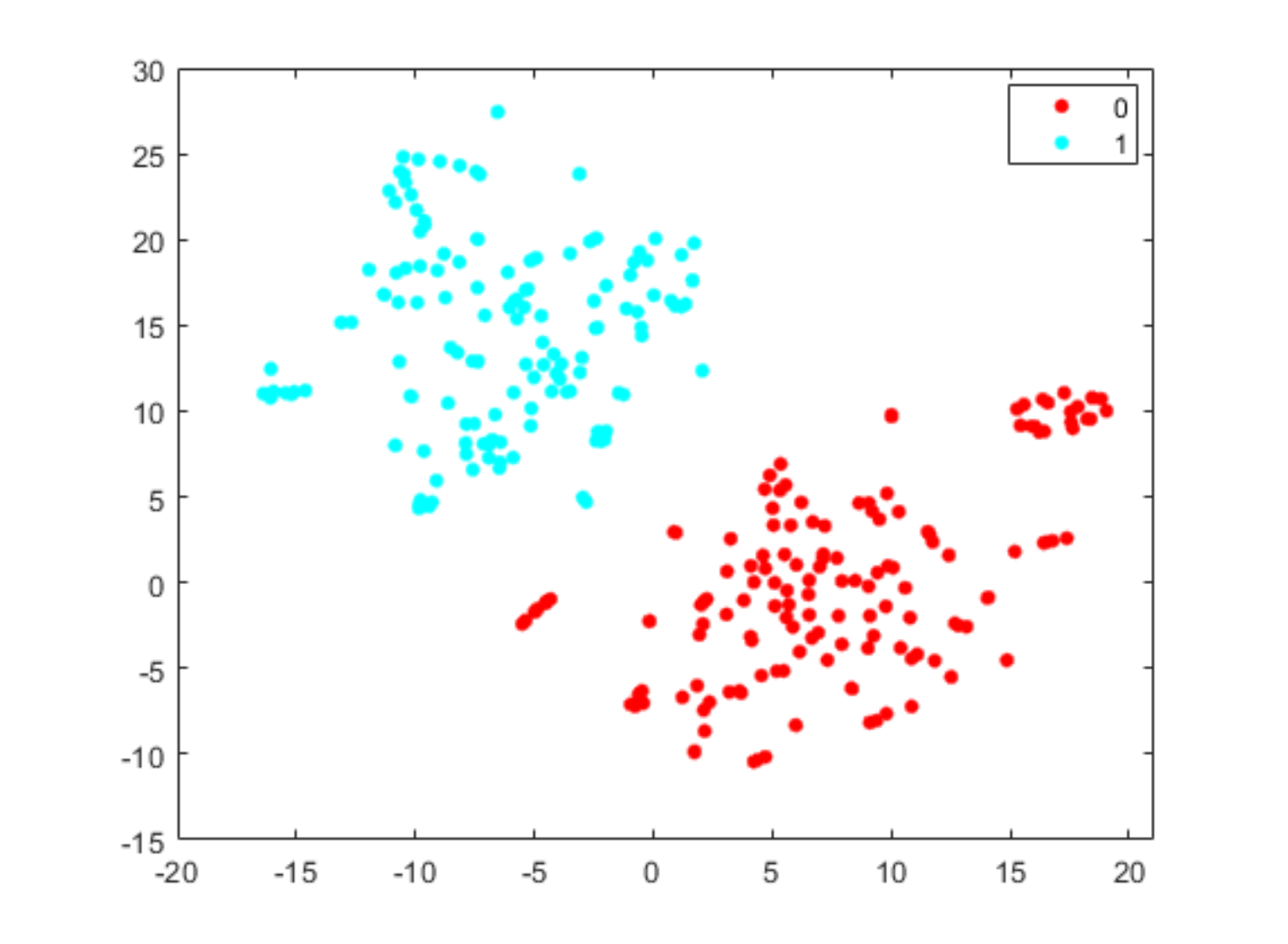}
  \centerline{(d)}
\end{subfigure}\hfil 
\begin{subfigure}{0.25\textwidth}
  \includegraphics[width=\linewidth]{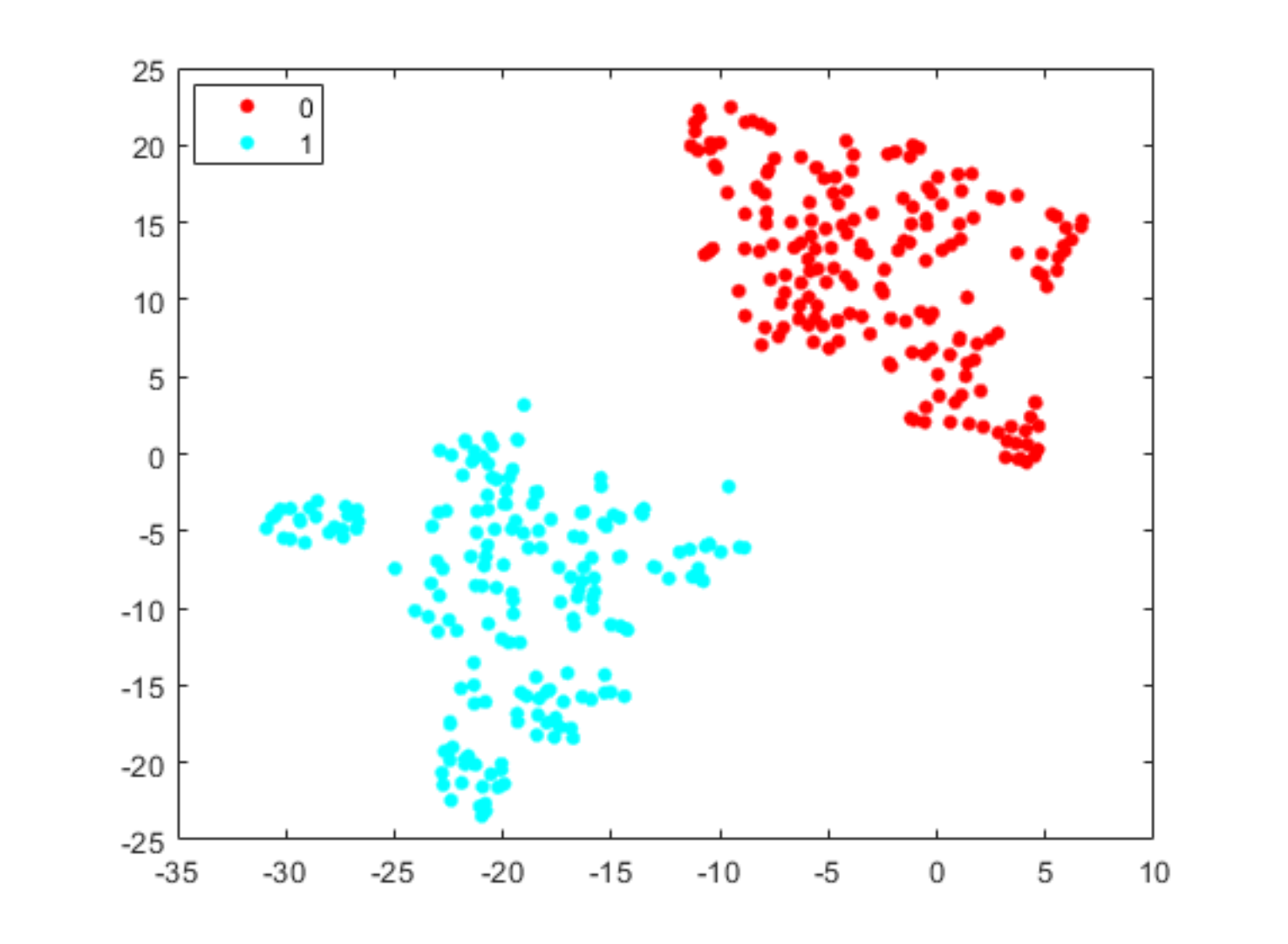}
  \centerline{(e)}
\end{subfigure}\hfil
\begin{subfigure}{0.25\textwidth}
  \includegraphics[width=\linewidth]{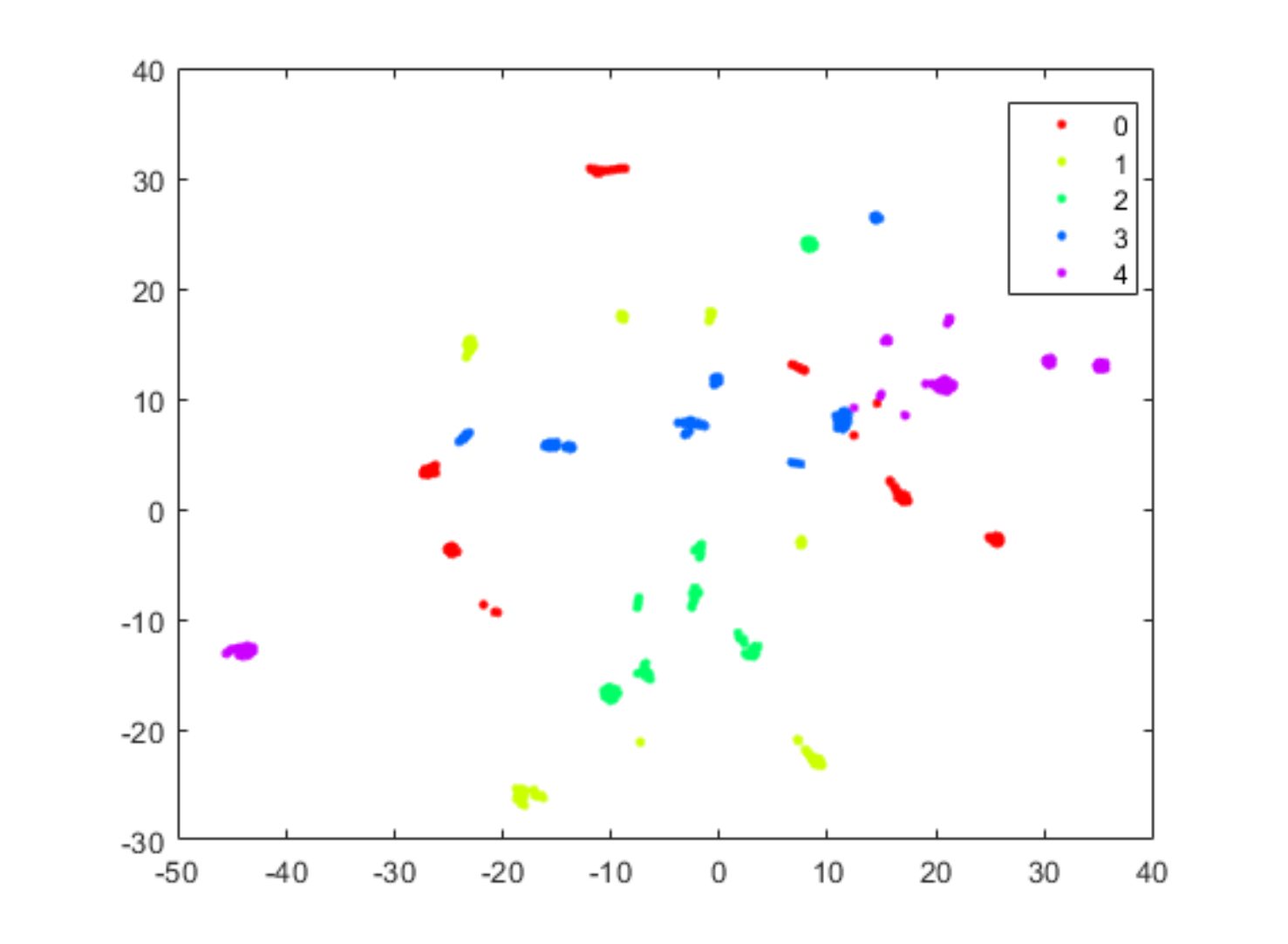}
  \centerline{(f)}
\end{subfigure}\hfil 
\begin{subfigure}{0.25\textwidth}
  \includegraphics[width=\linewidth]{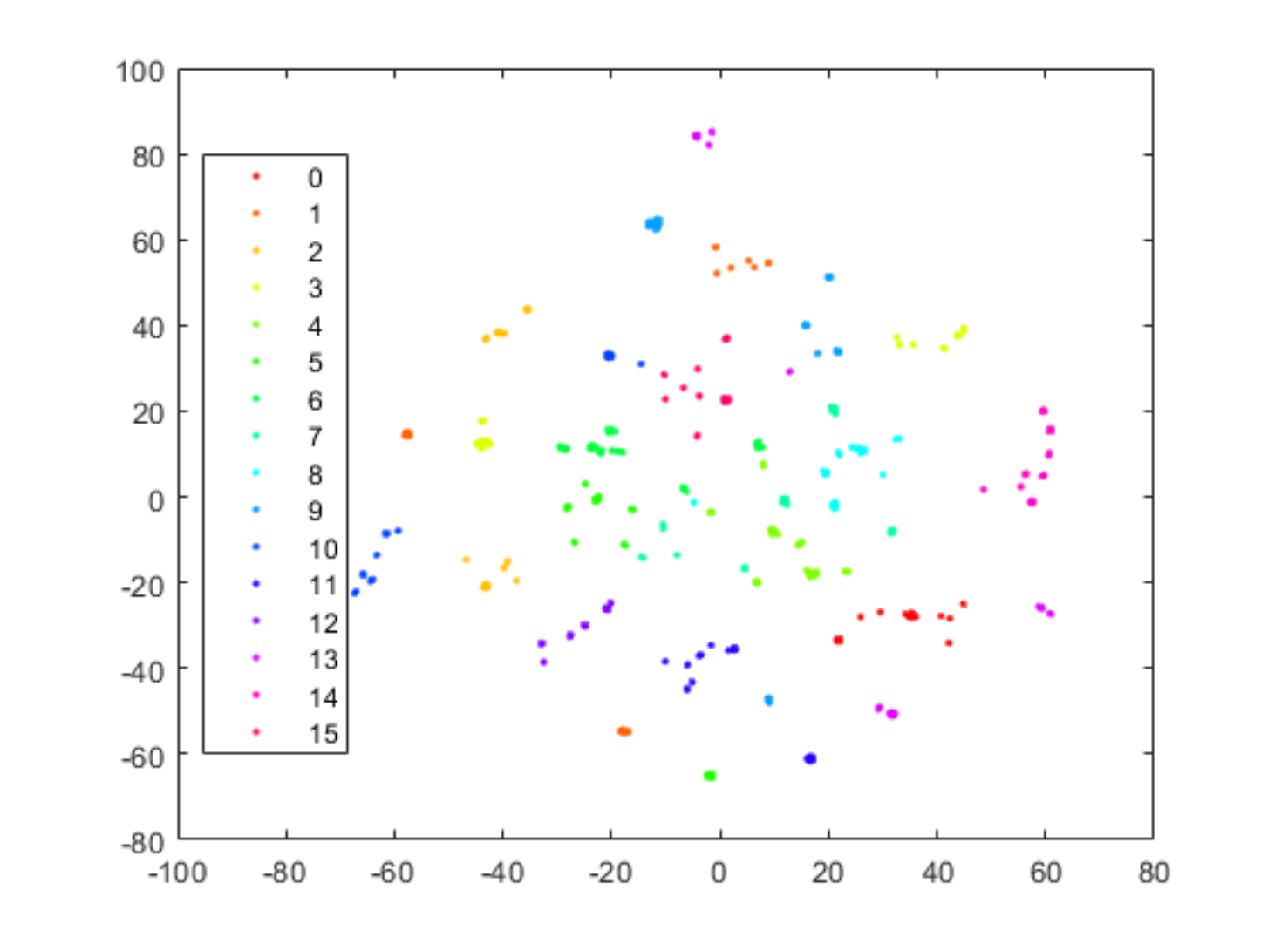}
  \centerline{(g)}
\end{subfigure} 
\caption{Results obtained with t-SNE on different datasets demonstrating the separation between different devices belonging to the same model: (a) two iPhone 5c (b) two iPhone 4s (c) two Fujifilm X-A10 (d) two Redmi Note 3 (e) two Asus Zenfone Max Pro M1 (f) five Xioami 4A and (g) 16 devices of 6 different models from the Daxing dataset.}
\label{fig:3}
\end{figure*}

\begin{table}[!t]
\centering
\caption{PCE evaluation for the PRNUs extracted from the original and down-sampled images w.r.t. $\hat{F}_{nat}$ and $\hat{F}_{flat}$}
\begin{tabular}{|c|c|c|c|}
\hline
\textbf{Device}                                                             & \textbf{\begin{tabular}[c]{@{}c@{}}Average PCE\\of PRNUs of\\30 original\\ images w.r.t.\\ $\hat{F}_{nat}$\end{tabular}} & \textbf{\begin{tabular}[c]{@{}c@{}}Average PCE of\\PRNUs of\\30 PRNU-free\\ images w.r.t.\\ $\hat{F}_{nat}$\end{tabular}} & \textbf{\begin{tabular}[c]{@{}c@{}}Average PCE of\\PRNUs of\\30 PRNU-free\\ images w.r.t.\\ $\hat{F}_{flat}$\end{tabular}} \\ \hline
\textbf{iPhone 5c-1}                                                        & 16,067                                                                                   & 1.52                                                                                           & 2.19                                                                                           \\ \hline
\textbf{iPhone 5c-2}                                                        & 9,848                                                                                    & 1.96                                                                                           & 2.93                                                                                           \\ \hline
\textbf{iPhone 4s-1}                                                        & 17,674                                                                                   & 1.22                                                                                           & 1.89                                                                                           \\ \hline
\textbf{\begin{tabular}[c]{@{}c@{}}Fujifilm\\ X-A10-1\end{tabular}}         & 1,095                                                                                    & 1.51                                                                                           & -                                                                                                \\ \hline
\textbf{\begin{tabular}[c]{@{}c@{}}Redmi\\ Note 3-1\end{tabular}}           & 14,845                                                                                   & 1.51                                                                                           & -                                                                                                \\ \hline
\end{tabular}
\label{table2}
\end{table}

\subsection{SDI Based on Down-Sampled Patches}
\label{resB}
For each camera, the reference PRNU ($\hat{F}_{nat}$) utilized during the PCE evaluation is estimated using 50 natural images. In the case of the VISION dataset, we also generate the reference PRNU ($\hat{F}_{flat}$) using 50 flat-field images. To evaluate the effectiveness of PRNU removal through image down sampling, we compare the average PCE values over 30 test images before and after removing the PRNU with the proposed PRNU removal method as detailed in Section \ref{methA}. From the results in Table \ref{table2}, we can see that there is a substantial reduction in the PCE values after removing the PRNU from the images. Be it the reference PRNU $\hat{F}_{nat}$ estimated from natural images or $\hat{F}_{flat}$ estimated from flat-field images, the average PCE values for the down-sampled images are significantly less than the detection threshold 50, indicating the effectiveness of the proposed down-sampling PRNU removal method.

We further present our results on the establishment of the presence of a new non-PRNU device-specific fingerprint. We have shown that down sampling is effective in removing PRNUs. Thus, to demonstrate that device-specific fingerprints can be effectively extracted from PRNU-free images, we first down-sample all images in the training and testing sets to 224$\times$224$\times$3 pixels and then train the ResNet101 model on these PRNU-free images from different devices of the same camera model. We carry out various experiments by training the ResNet101 model on different datasets. The devices used for each experiment are listed in Table \ref{table3}. The feature extracted from the GAP layer of the trained ResNet101 model is 2048-dimensional. To visualize the high-dimensional features, we use the t-distributed Stochastic Neighbor Embedding (t-SNE) and show the results in Fig. \ref{fig:3}. As we can see, the separation between the devices belonging to the same brand and model is well evidenced, which suggests that \emph{there is a unique device-specific fingerprint other than PRNU in the images}.

\begin{table}[!t]
\centering
\caption{SDI performance based on the non-PRNU fingerprint extracted from down-sampled and random-sampled images}
\label{table3}
\begin{tabular}{|c|c|c|c|c|}
\hline
\multirow{2}{*}{\textbf{\vtop{\hbox{\strut }\hbox{\strut }\hbox{\strut }\hbox{\strut Exp.}}}} & \multirow{2}{*}{\textbf{\vtop{\hbox{\strut }\hbox{\strut }\hbox{\strut }\hbox{\strut Device used}}}}  & \multicolumn{3}{c|}{\textbf{Testing Accuracy (\%)}} \\ \cline{3-5}
&           &\textbf{\vtop{\hbox{\strut Down}\hbox{\strut sampling}\hbox{\strut from}\hbox{\strut original}\hbox{\strut images}}} & \textbf{\vtop{\hbox{\strut Random}\hbox{\strut sampling}\hbox{\strut from}\hbox{\strut original}\hbox{\strut images}}} &\textbf{\vtop{\hbox{\strut Random}\hbox{\strut sampling}\hbox{\strut from down-}\hbox{\strut sampled}\hbox{\strut images}}}              \\ \hline
\textbf{1}          & \begin{tabular}[c]{@{}c@{}}2 iPhone 5c \end{tabular}                                                  & 97.06                  & 83.86     & 84.44    \\ \hline
\textbf{2}          & \begin{tabular}[c]{@{}c@{}}2 iPhone 4s\end{tabular}                                                  & 98.00                  & 85.70      & 86.38    \\ \hline
\textbf{3}          & \begin{tabular}[c]{@{}c@{}}2 Fujifilm X-A10\end{tabular}                                        & 95.92                 & 84.39      & 86.27     \\ \hline
\textbf{4}          & \begin{tabular}[c]{@{}c@{}}2 Redmi Note 3\end{tabular}                                            & 97.00               & 85.38      & 92.26      \\ \hline
\textbf{5}          & \begin{tabular}[c]{@{}c@{}}2 Asus Zenfone\\Max Pro M1\end{tabular}                  & 95.16                      & 84.25   & 86.24  \\ \hline
\textbf{6}          & 5 Xiaomi 4A                                                                                                & 99.57          & 94.01             & 95.79   \\ \hline
\textbf{7}          & \begin{tabular}[c]{@{}c@{}}16 devices \\of 6 different \\models from the \\Daxing dataset\end{tabular}                           & 99.20            & 98.15          & 99.58                 \\ \hline
\end{tabular}
\end{table}

\begin{figure}[!t]
\centering
\includegraphics[width=3.55in]{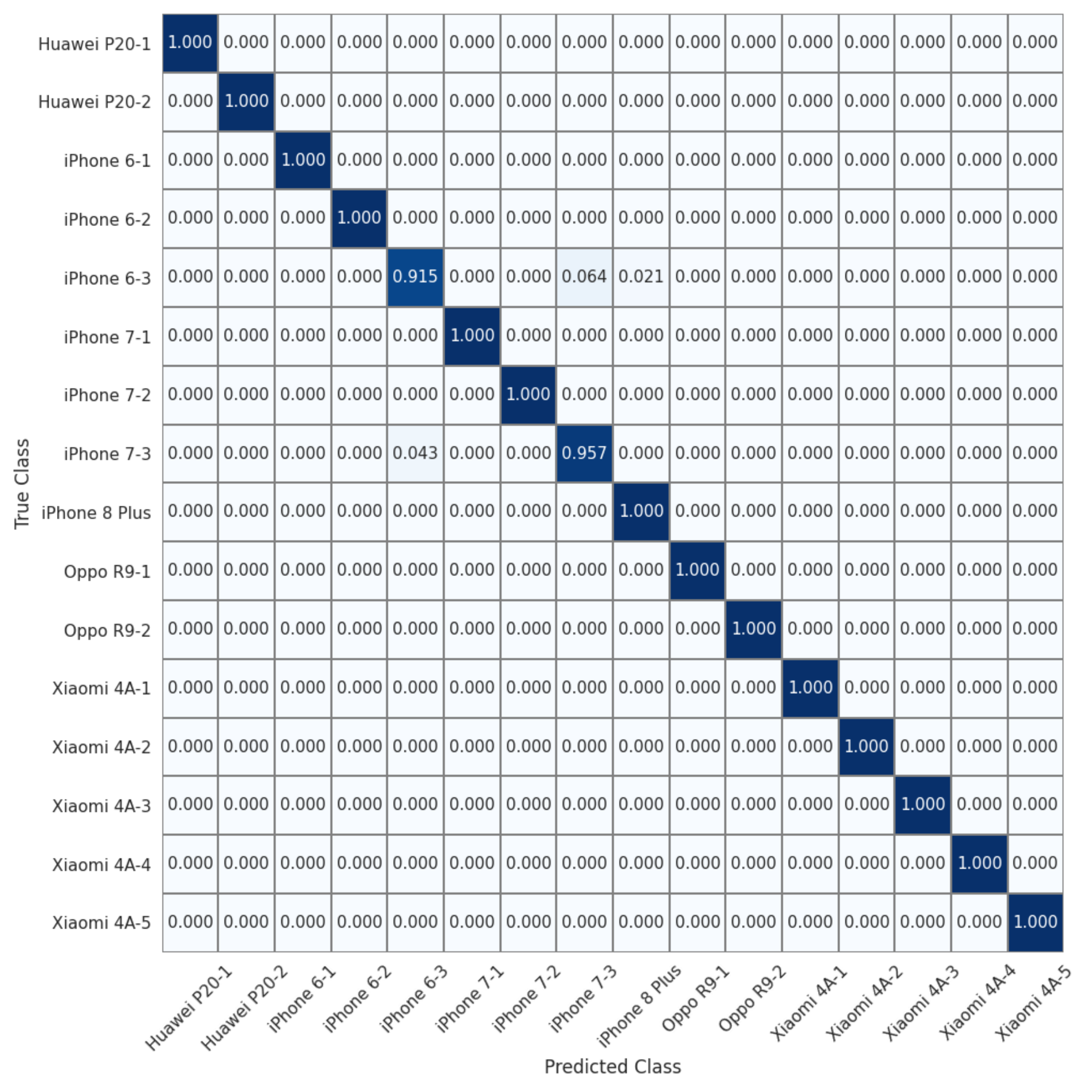}
\caption{Normalized confusion matrix for SDI based on down-sampled images extracted from original images for 16 devices of 6 different models in the Daxing dataset.}
\label{fig4}
\end{figure}

\par Furthermore, the down sampling operation removes the high-frequency components in the images, so only the low and mid-frequency components of the images were taken into account by the ResNet101 model while learning the non-PRNU device fingerprint. This indicates that the features that serve as \emph{the device-specific fingerprint extracted by the proposed model are mainly embedded in the low and mid-frequency components of the image}. Since high-frequency components of the image are sensitive to manipulation and counter forensic attacks, we cannot rely on the device fingerprints concealed in the high-frequency band for forensic purposes. However, the proposed method makes evident the presence of the new device fingerprint in the low- and mid-frequency band that can be used as the robust fingerprint to identify the exact source device of the image.

\par Similar to the work in \cite{bondi}, we train an SVM classifier on the 2048-dimensional features extracted by the ResNet101 model to test their capability in identifying the exact source device of the image. The performance of the SVM classifier on exact device identification is evaluated in terms of classification accuracy and is reported in the third column of Table \ref{table3}. It can be observed that the proposed model is able to map the PRNU-free images to their respective source cameras accurately based on the device-specific fingerprint extracted by the ResNet101 model. Furthermore, the proposed model achieves the best testing accuracy of 99.57\% and 99.20\% for Experiment 6 and Experiment 7, respectively. Note that the 5 devices of the Xiaomi 4A model in Experiment 6 and the 16 devices of 6 different models in Experiment 7 are from the Daxing dataset, which includes flat-field images such as the images of clear sky and white wall that are uniformly illuminated. As flat-field images have less intensity variation, the proposed deep learning model is less susceptible to scene details. This makes the model focus more on the intrinsic information of the images which contributes to learning the fingerprint. As a result, the proposed model achieved higher accuracy on the Daxing dataset compared to the other datasets.

\begin{figure}[!t]
\centering
\includegraphics[width=3.55in]{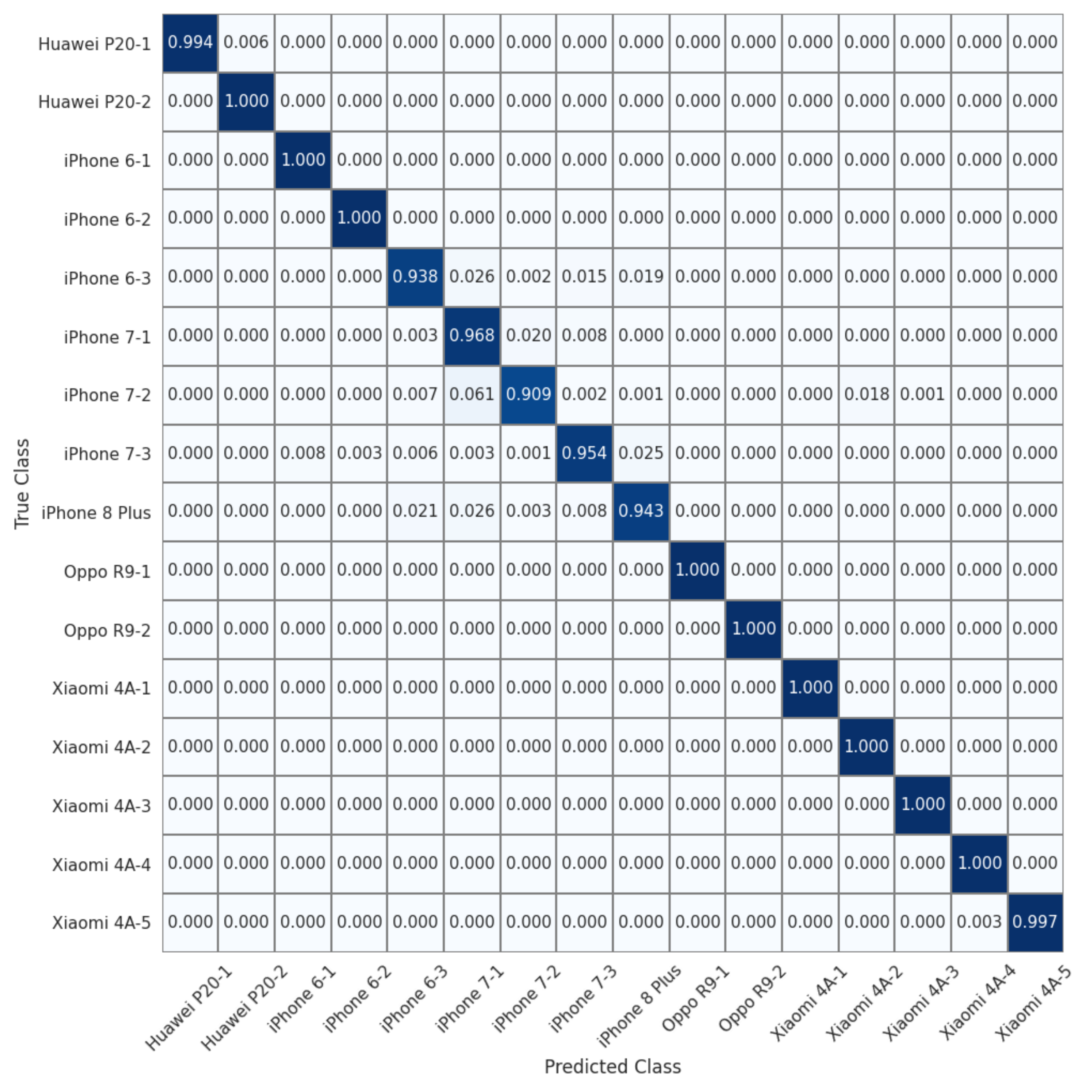}
\caption{Normalized confusion matrix for SDI based on random-sampled images extracted from original images for 16 devices of 6 different models in the Daxing dataset.}
\label{fig5}
\end{figure}

\begin{figure}[!t]
\centering
\includegraphics[width=3.55in]{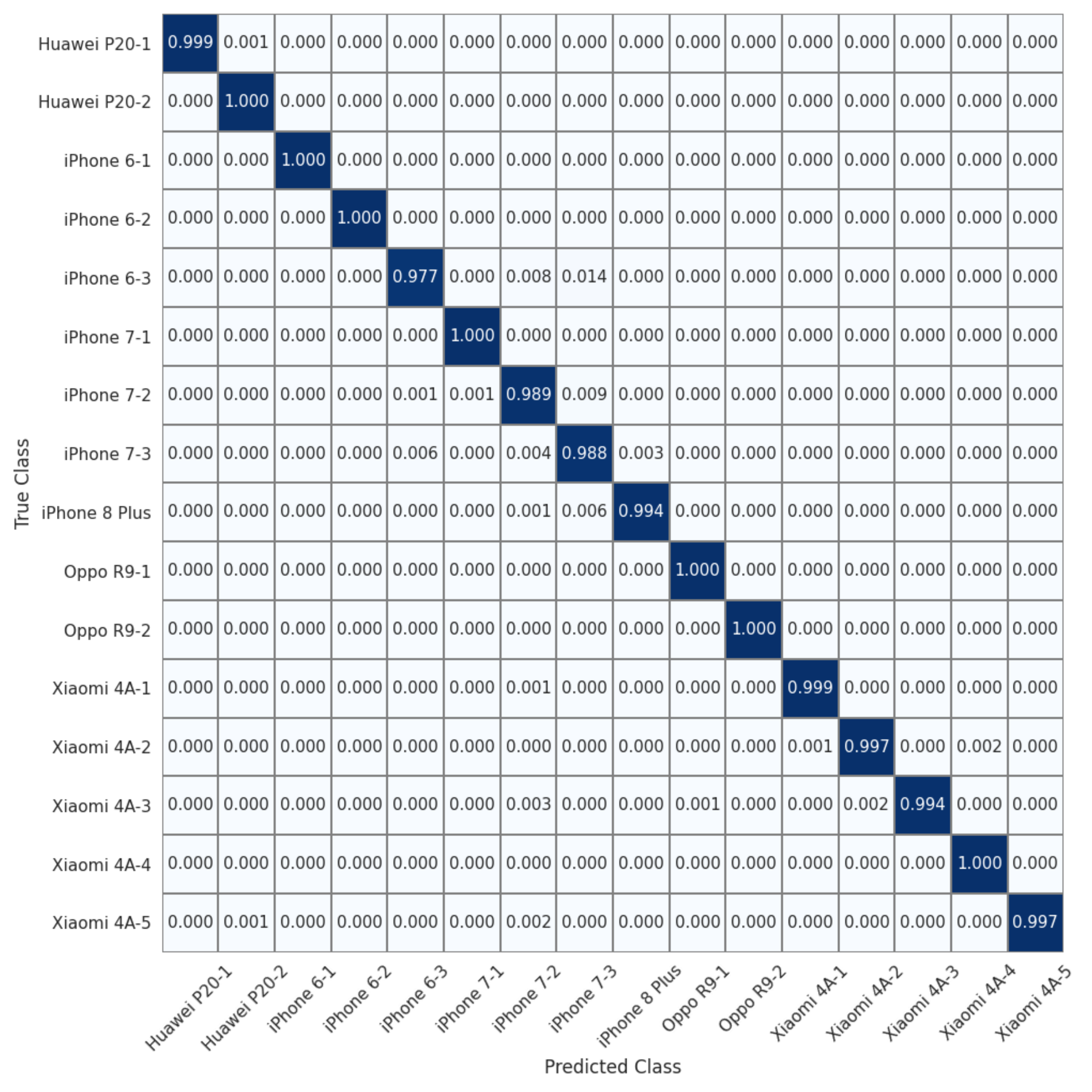}
\caption{Normalized confusion matrix for SDI based on random-sampled patches extracted from down-sampled images for 16 devices of 6 different models from the Daxing dataset.}
\label{fig6}
\end{figure}

\subsection{SDI Based on Random-Sampled Patches}
\label{resD}
In this subsection, we evaluate the SDI performance of the new device-specific fingerprint extracted from random-sampled patches, as detailed in Section \ref{methD}. To study whether or not there are global stochastic device-specific characteristics that can act as device fingerprints, from every image in the training and testing sets, we extract 50 patches with pixels taken at random from the original images or down-sampled images. The proposed hybrid ResNet-SVM system is trained on these random-sampled patches to extract the non-PRNU device fingerprint and identify the source device. We repeat the same experiments in the previous subsection for the proposed random sampling method and report the results in the fourth and fifth columns of Table \ref{table3}. Compared to the results of down sampling, the SDI accuracy of random sampling drops notably from about $95\%$ to $85\%$ for the cases of two devices. This is not surprising as the local structural information of the original image is abandoned for training. However, what is most striking is that the identification accuracy ($>83\%$) of random sampling are much higher than that of a random guess. Furthermore, the influence of the absence of structural information tends to diminish as the number of devices increases in Experiment 6 and Experiment 7. 

To allow for a more detailed comparison, we show the normalized confusion matrices on the 16 devices of the Daxing dataset (Experiment 7) in Fig. \ref{fig4}-\ref{fig6}. We can see that the SDI performance only drops slightly for a few iPhone devices, i.e. iPhone 7 and iPhone 8 Plus. These results appear to indicate that involving more devices in the training is beneficial for learning discriminative device-specific fingerprint. However, as we involve more devices in the training, we should not overlook the potential chance of learning model-specific rather than device-specific fingerprint, because the number of camera models is usually larger than the average number of devices of the same model in a dataset. Thus, further studies are required to investigate the appropriate way of dataset construction for learning discriminative device-level fingerprint. Another interesting observation from Table \ref{table3} is that random sampling from the down-sampled images slightly outperforms random sampling from the original images. We contend that most high-frequency components removed by down sampling can be considered as interference for device identification. Therefore, similar to the common practice of suppressing the scene details to improve the performance of PRNU-based source device identification \cite{Li}, random sampling from the down-sampled images rules out the effect of interferential high-frequency components in the images and thus improves the device identification performance.

For random sampling, each patch is formed with pixels taken at random from the image, the spatial relationship among the pixels within each of the patches is disturbed. Therefore, no structural information (e.g., the model-specific periodic artifacts due to JPEG compression or demosaicing) is maintained in each patch. The results of random sampling in Table \ref{table3} suggest the existence of stochastic and location-independent device-specific characteristics that can be captured by random sampling to serve as device fingerprint. The PRNU fingerprint, being a deterministic component, requires proper spatial synchronization, post-processing such as resizing, cropping, and filtering of the images prevent authentic source identification based on PRNU. Thus, relying on the fragile PRNU for forensic purposes will mislead the investigations. With the new characteristics being stochastic and globally present, it opens the opportunity to learn robust device-specific fingerprint that is less susceptible to the desynchronization operations and counter-forensic attacks on the images.

\subsection{Performance Comparison with State-of-the-Art Techniques}
\label{resE}
The existing literature on source camera identification \cite{bondi, tuama, yao, fre, huang, wang, chen2017,yang2019,ding2019,sameer2020,mandelli} has shown excellent performance on camera model identification. However, our preliminary analysis showed that the state-of-the-art techniques on camera model identification cannot be used directly to identify the exact device. Most, if not all, existing data-driven approaches \cite{bondi, yao,fre,huang,wang,yang2019} report a drastic reduction in the device-level classification accuracy when PRNU-free images are used for training. In this subsection, we compare the performance of the proposed hybrid ResNet-SVM system with the state-of-the-art data-driven approaches. We train the state-of-the-art techniques using PRNU-free images generated through the down sampling operation. The classification accuracy achieved on the testing set of different datasets is reported in Table \ref{table4}. It can be observed that in the case of individual device identification, both the proposed down sampling and random sampling methods outperform the existing data-driven approaches by a large margin. The content-adaptive fusion network \cite{yang2019} developed for individual device identification exhibits a low classification accuracy. Also, the CNN proposed in \cite{wang} that learns the intrinsic source information (such as PRNU, lens distortion noise pattern and traces of color dependencies related to CFA interpolation) through the local binary pattern (LBP) pre-processing layer, shows a poor performance on PRNU-free images. Thus, their results indicate that the PRNU remains an important characteristic for the existing techniques to identify the source camera. They only work when a strong PRNU fingerprint is present in the image. Nonetheless, it can be seen that \emph{our hybrid ResNet-SVM system is capable of identifying the individual cameras of the same model from the PRNU-free down-sampled and random-sampled image patches}. 

We summarize two possible reasons that contribute to the superior performance of our proposed method: 
1) Deeper networks allow for learning more discriminative features for SDI. However, some existing works, e.g. \cite{bondi,yao,fre,huang,wang,yang2019}, used shallow neural networks (less than 15 convolutional layers) to learn features, which might not be sufficiently discriminative for device-level source camera identification. 2) Probably inspired by the success of PRNU, it is generally believed that the intrinsic source device information largely resides in the noise residual or the high-frequency domain of an image. For this reason, most existing data-driven approaches \cite{cozzolino2018noiseprint,zuo2018camera,quan2020provenance,tuama,wang} attempt to learn discriminative features from the noise residuals or high-pass filtered images rather than the original images. In contrast to these approaches, our proposed methods successfully learn the device-specific fingerprint from PRNU-free images, which indicates that PRNU noise may only be one type, or make up a small part, of device-specific fingerprints.

\begin{table}[!t]
\centering
\caption{Performance comparison of proposed methods with state-of-the-art techniques.}
\label{table4}
\begin{tabular}{|c|c|c|c|c|}
\hline
\multirow{2}{*}{\textbf{Method}}                                                 & \multicolumn{4}{c|}{\textbf{Testing Accuracy (\%)}}                                                                                                                                                                                                                                           \\ \cline{2-5} 
                                                                                 & \textbf{\begin{tabular}[c]{@{}c@{}}2\\ iPhone 4s\end{tabular}} & \textbf{\begin{tabular}[c]{@{}c@{}}2\\ Fujifilm\\ X-A10\end{tabular}} & \textbf{\begin{tabular}[c]{@{}c@{}}2\\ Redmi\\ Note 3\end{tabular}} & \textbf{\begin{tabular}[c]{@{}c@{}}16 devices\\ (Daxing\\ dataset)\end{tabular}} \\ \hline
\textbf{\begin{tabular}[c]{@{}c@{}}CNN+SVM \cite{bondi}\end{tabular}}          & 50.00                                                          & 50.00                                                               & 50.00                                                               & 6.25                                                                             \\ \hline
\textbf{\begin{tabular}[c]{@{}c@{}}CNN based multi-\\classifier in \cite{yao}\end{tabular}}            & 76.00                                                          & 58.16                                                               & 75.00                                                               & 76.60                                                                            \\ \hline
\textbf{\begin{tabular}[c]{@{}c@{}}CNN with 6 layers\\ \cite{fre}\end{tabular}} & 73.00                                                          & 45.00                                                               & 76.00                                                               & 21.15                                                                            \\ \hline
\textbf{\begin{tabular}[c]{@{}c@{}}CNN with 11 layers\\ \cite{huang}\end{tabular}}          & 50.00                                                          & 50.00                                                               & 50.00                                                               & 6.25                                                                             \\ \hline
\textbf{\begin{tabular}[c]{@{}c@{}}LBP+modified\\AlexNet \cite{wang}\end{tabular}}           & 64.71                                                          & 50.00                                                               & 53.00                                                               & 6.25                                                                            \\ \hline
\textbf{\begin{tabular}[c]{@{}c@{}}Content-adaptive\\fusion network \cite{yang2019}\end{tabular}}           & 70.00                                                          & 59.00                                                               & 62.73                                                               & 51.06                                                                            \\ \hline
\textbf{\begin{tabular}[c]{@{}c@{}}Proposed down\\ sampling from \\original images \end{tabular}}               & \textbf{98.00}                                                 & \textbf{95.92}                                                      & \textbf{97.00}                                                      & \textbf{99.20}                                                                   \\ \hline
\textbf{\begin{tabular}[c]{@{}c@{}}Proposed random\\ sampling from \\original images \end{tabular}}               & \textbf{85.70}                                                 & \textbf{84.39}                                                      & \textbf{85.38}                                                      & \textbf{98.15} \\ \hline
\textbf{\begin{tabular}[c]{@{}c@{}}Proposed random\\ sampling from \\down-sampled images \end{tabular}}               & \textbf{86.38}                                                 & \textbf{86.27}                                                      & \textbf{92.26}                                                      & \textbf{99.58} \\ \hline
\end{tabular}
\end{table}

\begin{table*}[!t]
\centering
\caption{Identification accuracy (\%) of the proposed method under various image manipulations}
\label{table5}
\begin{tabular}{|c|c|c|c|c|c|c|c|c|c|}
\hline
\multirow{2}{*}{\textbf{Device used for the experiment}}                                         & \multirow{2}{*}{\textbf{Original}} & \multicolumn{2}{c|}{\textbf{Gamma Correction}} & \multicolumn{3}{c|}{\textbf{Rotation}}  & \multicolumn{3}{c|}{\textbf{JPEG Compression}}      \\ \cline{3-10} 
                                                                                                 &                                    & $\gamma = 0.7$       & $\gamma = 1.4$      & $\theta=15^{\circ}$ & $\theta=30^{\circ}$ & $\theta=90^{\circ}$ & $Q = 90\%$ & $Q = 50\%$ & $Q = 20\%$ \\ \hline
\textbf{2 iPhone 5c devices}                                                                     & 97.06                              & 97.06                  & 96.08                 & 96.08       & 87.25       & 91.18       & 97.06           & 96.08           & 96.08           \\ \hline
\textbf{2 Fujifilm X-A10 devices}                                                                & 95.92                              & 95.92                  & 94.95                 & 91.84       & 88.78       & 85.71       & 95.92           & 95.92           & 95.92           \\ \hline
\textbf{2 Redmi Note 3 devices} & 97.00                              & 97.00                  & 94.00                 & 97.00       & 94.00       & 83.00       & 97.00           & 97.00           & 96.00           \\ \hline
\end{tabular}
\end{table*}

\subsection{Evaluation of Robustness Against Image Manipulations}
\label{resF}
In practical circumstances, the images are often subjected to various manipulations, which requires the source identification techniques to be robust against these manipulations. In this subsection, we test the robustness of the proposed method against various image manipulations such as gamma correction (with $\gamma {\in} \{0.7, 1.4\}$), rotation (with rotation angle $\theta{\in}\{15^{\circ}, 30^{\circ}, 90^{\circ}\}$), and JPEG compression (with quality factor $Q {\in} \{90\%, 50\%, 20\%\}$), thereby evaluate the capability of the new device fingerprint. To verify the robustness we have considered the proposed system trained on the original down-sampled images (i.e. no manipulations applied). For the testing process, all the images in the test set are processed with the aforementioned image manipulations and then down-sampled to remove PRNU fingerprint.
\par The SDI performance of the proposed method for various image manipulations on different cameras are presented in Table \ref{table5}. It can be seen that the proposed method is effective in identifying the images subjected to gamma correction. The identification accuracy on the rotated images is slightly lower than that on the original images. JPEG compression is widely used on social networks which suppresses the PRNU fingerprint left on the high-frequency band of the images making source identification a difficult task. However, it is evident from the results that the proposed method is almost unaffected by the JPEG compression attack on the images. This is not surprising because the new device-specific fingerprint resides in the low and mid frequency band, so even if the JPEG compression removes the high-frequency components in the image, the proposed method is still effective to identify the source device of the images.

\section{Conclusion}
\label{con}
In this paper, we explore the existence of a new device-specific fingerprint other than the PRNU in the low and mid-frequency bands of images for source device identification. The presence of the new device fingerprint in the low and mid-frequency bands of images is established by training a hybrid ResNet101-SVM model on down-sampled and random-sampled image patches. While the existing data-driven approaches can differentiate only camera brands or models, the experimental results suggest that the proposed system is effective in extracting the non-PRNU device-specific fingerprint and mapping the images to the individual devices of the same model. The source device identification based on the new device-specific fingerprint presented in this paper provides significantly more accurate and reliable results compared to earlier data-driven approaches that depend on the PRNU fingerprint. Moreover, the PRNU-based approaches demand proper spatial alignment of the images to identify the source camera. Nonetheless, the location independent and global stochastic characteristics of our new device fingerprint that learned from random-sampled patches make it possible to identify the source camera even with the presence of spatial misalignment of the images. We have also investigated the reliability of the new fingerprint from images subjected to gamma correction, rotation and JPEG compression. It shows that the new fingerprint is effective and robust in identifying the source camera of the images subjected to aggressive JPEG compression. With these promising results, we establish the presence of a robust device-specific fingerprint that can be used for image forensic applications.


%

%
%
%
%
%


\ifCLASSOPTIONcaptionsoff
  \newpage
\fi



\bibliographystyle{IEEEtran}
\bibliography{manuscript}
\end{document}